**RESEARCH**

# Enhancing ASD detection accuracy: a combined approach of machine learning and deep learning models with natural language processing

Sergio Rubio-Martín[1*†] , María Teresa García-Ordás[2†], Martín Bayón-Gutiérrez[2†], Natalia Prieto-Fernández[2†] and José Alberto Benítez-Andrades[1†]

**Abstract**

**Purpose:** The main aim of our study was to explore the utility of artificial intelligence (AI) in diagnosing autism spectrum disorder (ASD). The study primarily focused on using machine learning (ML) and deep learning (DL) models to detect ASD potential cases by analyzing text inputs, especially from social media platforms like Twitter. This is to overcome the ongoing challenges in ASD diagnosis, such as the requirement for specialized professionals and extensive resources. Timely identification, particularly in children, is essential to provide immediate intervention and support, thereby improving the quality of life for affected individuals.

**Methods:** We employed natural language processing (NLP) techniques along with ML models like decision trees, extreme gradient boosting (XGB), k-nearest neighbors algorithm (KNN), and DL models such as recurrent neural networks (RNN), long short-term memory (LSTM), bidirectional long short-term memory (Bi-LSTM), bidirectional encoder representations from transformers (BERT and BERTweet). We extracted a dataset of 404,627 tweets from Twitter users using the platform's API and classified them based on whether they were written by individuals claiming to have ASD (ASD users) or by those without ASD (non-ASD users). From this dataset, we used a subset of 90,000 tweets (45,000 from each classification group) for the training and testing of these models.

**Results:** The application of our AI models yielded promising results, with the predictive model reaching an accuracy of almost 88% when classifying texts that potentially originated from individuals with ASD.

**Conclusion:** Our research demonstrated the potential of using AI, particularly DL models, in enhancing the accuracy of ASD detection and diagnosis. This innovative approach signifies the critical role AI can play in advancing early diagnostic techniques, enabling better patient outcomes and underlining the importance of early identification of ASD, especially in children.

†Sergio Rubio-Martín, María Teresa García-Ordás, Martín Bayón-Gutiérrez, Natalia Prieto-Fernández, and José Alberto Benítez-Andrades have contributed equally to this work.

*Correspondence:  srubm@unileon.es
[1] SALBIS Research Group, Dept. of Electric, Systems and Automatics Engineering, Universidad de León, Campus de Vegazana s/n, 24071 León, León, Spain
Full list of author information is available at the end of the article

## Introduction

Autism spectrum disorder (ASD) is a developmental disability that impacts individuals' social and interactive skills when engaging with others [1]. The condition typically manifests before the age of three and can persist throughout a person's life, leading to a lower quality of life for those who remain undiagnosed in childhood [2]. ASD encompasses a wide range of subtype conditions, with one of the subtypes known as Asperger Syndrome (AS), which is classified as severity 1 within the autism spectrum [3]. AS was officially recognized as an ASD subtype in 2013 by the Diagnostic and Statistical Manual of Mental Disorders (DSM-5) and later by the World Health Organization (WHO) in its International Classification of Diseases 11th Revision (ICD-11) in 2022 [3].

Early diagnosis of ASD offers numerous scientific advantages. Firstly, early intervention significantly improves social, communicative, and cognitive skills in children with ASD [4, 5]. Detecting the disorder at an early age allows professionals to implement tailored interventions that address the specific needs of the child, leading to better long-term development. Secondly, early diagnosis provides parents and caregivers with access to appropriate resources and support, improving the overall quality of life for both the child and the family [6]. This includes specialized therapies, counseling, and adjustments to the child's environment to accommodate their unique requirements. Additionally, early ASD diagnosis positively impacts the individual's educational and occupational trajectory, offering increased opportunities for success in academic and work settings [7]. By providing suitable interventions from the outset, children have greater chances to acquire skills necessary for future achievements. Lastly, early diagnosis contributes to scientific research by enhancing our understanding of the causes, progression, and variability of ASD. This knowledge aids in identifying risk factors and developing more effective treatment and prevention strategies [8].

While many studies have focused on detecting ASD, most rely on time-consuming interviews and questionnaires, emphasizing the need for early diagnosis [9]. However, recent advancements in artificial intelligence (AI) offer promising solutions in the healthcare field [10, 11]. AI models have been successfully applied to detect and diagnose various types of cancer, such as prostate, breast, and cervical cancer [12–14]. In the context of ASD, previous research has primarily utilized machine learning techniques and datasets consisting of images [15, 16], or focused on specific techniques like eye-tracking [17]. However, recent studies have explored the use of AI and machine learning to analyze textual data from social media platforms, particularly Twitter, to identify ASD-related behaviors and patterns [18]. By examining the content and structure of tweets, AI algorithms can provide valuable insights for early diagnosis and intervention, offering new opportunities for accurate and timely ASD diagnosis [19, 20].

Given the significance of social networks and AI in early ASD diagnosis, there is a research gap in utilizing information from Twitter users' biographies to develop models that aid in this endeavor [21]. Individuals within the ASD spectrum often disclose their condition in their biographies using hashtags, plain text, emojis, or emoticons, enabling more precise identification of individuals with different ASD subtypes. Twitter, with its accessibility for extracting textual data, is the platform of choice for this study. The primary objective is to develop artificial intelligence models that can diagnose ASD by analyzing the texts posted by users who openly disclose their condition in their Twitter biographies.

Building upon our preliminary study presented at a conference, where we delved into the potential of artificial intelligence for diagnosing Autism Spectrum Disorder (ASD) [22], this manuscript introduces several substantial advancements. Our key contributions in this extended research are:

- Additional Machine Learning Models: Beyond the models explored in our initial study, we have trained and evaluated others, notably including KNN. The rigorous process of including and evaluating these models, each with its unique characteristics and parameters, demanded significant effort for fine-tuning and optimization tailored to our dataset.
- Extended Deep Learning Models: We have further ventured into deep learning, incorporating models like RNN and LSTM, renowned for their prowess in handling data sequences like texts. This exploration also involved experimenting with various configurations to hone their performance, necessitating considerable computational resources and time.
- Pretrained BERT Models: Our exploration did not stop at conventional models. We ventured into BERT models, testing two distinct pretrained versions, each with its unique training datasets and features, influencing their performance on our tasks.

The magnitude of effort invested in testing these models with our dataset is immense. Each model underwent its distinct tuning, training, and validation process, demanding intensive computational resources and time. This rigorous approach was pivotal to pinpoint the model or combination thereof that yielded the best accuracy and performance for our specific problem.

These contributions not only expand the scope and depth of our initial study but also underscore our meticulous exploration of configurations and parameters. The overarching goal remained consistent: to enhance the accuracy and efficacy of our models in detecting ASD.

Furthering our contributions, this manuscript also introduces:

- A comprehensive Twitter dataset comprising 404,627 tweets from 252 distinct users, with 221 users explicitly indicating ASD in their biographies.
- The application of a diverse array of ML and DL techniques to forge predictive models, aiding in diagnosing patients based on discernible patterns in texts.
- A comparative analysis of the accuracy achieved by traditional ML techniques vis-á-vis DL techniques in the predictive models.

In summation, recognizing the burgeoning significance of social networks coupled with the prowess of artificial intelligence in early ASD diagnosis, it is imperative to harness the insights from Twitter users' biographies. By scrutinizing textual data and leveraging machine learning techniques, we can craft models that significantly aid in the precise and timely diagnosis of ASD. Such strides hold the promise to deepen our understanding of ASD-related behaviors and experiences, refine interventions, and ultimately uplift the lives of individuals with ASD and their families.

The structure of the paper is as follows: "Material and methods" section offers a comprehensive explanation of the methodology utilized in the various techniques proposed. In "Experiments and results" section, the experiments and results are presented, along with a comparative analysis of the different techniques. Finally, "Discussion and conclusions" section encompasses the discussion, where the conclusion is also presented for a cohesive narrative within the same section.

## Material and methods

The paper provides a detailed explanation of the research methodology in the following subsections. Firstly, "ASD dataset collection and classification" section describes the approach taken to obtain the complete dataset and outlines the classification process for each example. Moving on to "Machine learning and deep learning models used" section, the paper presents the machine learning and deep learning models employed to address the problem at hand. Additionally, "Hardware and Software used for the experiments" section provides an overview of the hardware specifications of the computer utilized in the research. The research outline can be visualized in Fig. 1.

### ASD dataset collection and classification

Initially, several datasets pertaining to ASD were explored, but they lacked sufficient representativeness. This lack of representativeness was primarily due to two factors: firstly, these datasets did not contain a sufficient number of records to train robust artificial intelligence models and secondly, the fields or columns within these datasets did not carry relevant information that could effectively contribute to the learning process of these models. Consequently, the decision was made to create a new dataset from scratch. To accomplish this, Twitter was chosen as the source of data, specifically focusing on English tweets from users who self-identified as having ASD in their biography profiles. The dataset extraction process involved programming a Python script capable of accessing the publicly available user data. Accessing this information required utilizing Twitter's API, which is exclusively accessible to developers who have undergone prior verification by the platform. Leveraging the API access, the data was exported to a CSV format for convenient handling and analysis. The configuration of the tweet data within the dataset was as follows:

- The initial step involved manual scraping of users by examining their biography profiles to determine whether they self-identified as having ASD. To identify these users, specific keywords such as 'Autism', 'ASD', 'Asperger', 'Aspie', 'Autistic', and 'ActuallyAutistic' were utilized. The inclusion of 'Asperger' and 'Aspie' as keywords stems from the recognition that Asperger's is now considered a subtype within the autism spectrum by the scientific community. Figure 2 illustrates two examples of ASD users, with certain data points such as username, location, and date of birth removed in compliance with Twitter's policies.

- During the user search process, each individual profile was meticulously reviewed to ensure accurate classification. It was important to discern between profiles belonging to individuals and those associated with organizations or societies, which led to the exclusion of certain profiles despite containing the relevant keywords. Additionally, there were cases where users were part of an ASD person's family, indicated by phrases such as 'father of an ASD kid' or 'mother of an ASD kid', which resulted in their exclusion from the dataset. Furthermore, some users identified themselves as 'ASD advocates', indicating their support for individuals with ASD but not personally having ASD themselves.

- Subsequently, the complete dataset consisting of tweets was automatically labeled by the programmer, considering the information available in the user's biography. As a result, two distinct groups were formed:
  - Tweets authored by users or individuals with ASD.
  - Tweets authored by users or individuals without ASD.

- Once we have a dataset composed of texts and a binary classification (written by individuals with ASD or without ASD), various artificial intelligence models are trained. These models take texts as input and output a classification, indicating whether the text was written by someone with ASD or not.

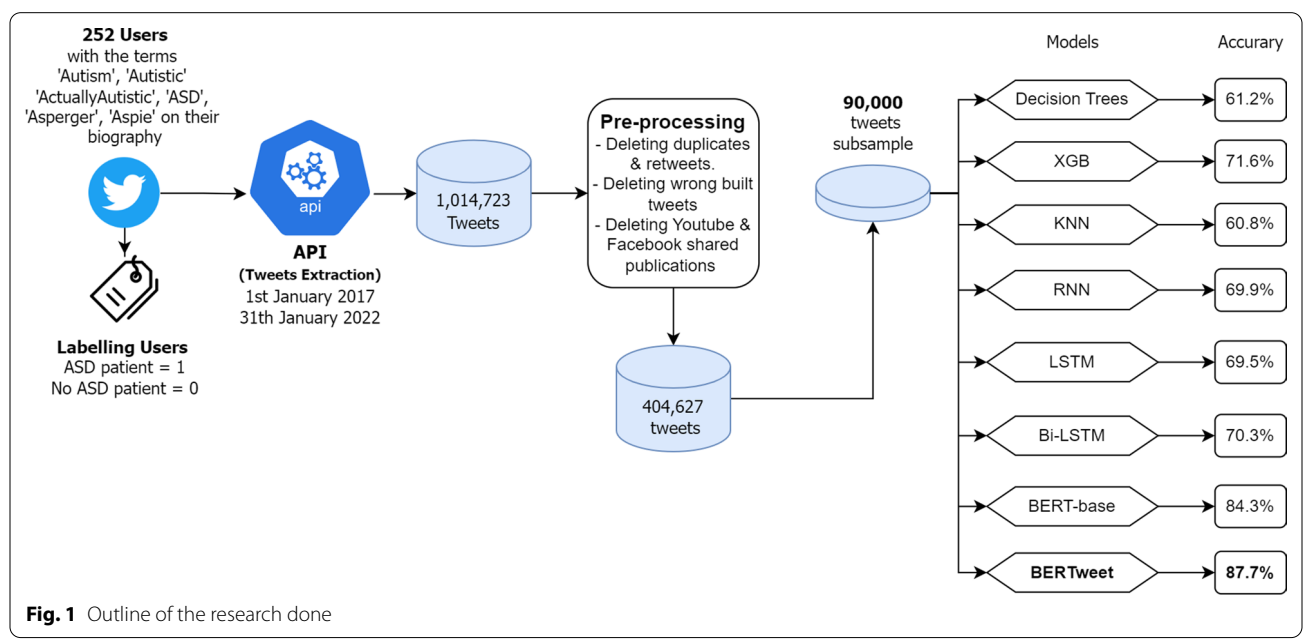

**Fig. 1** Outline of the research done

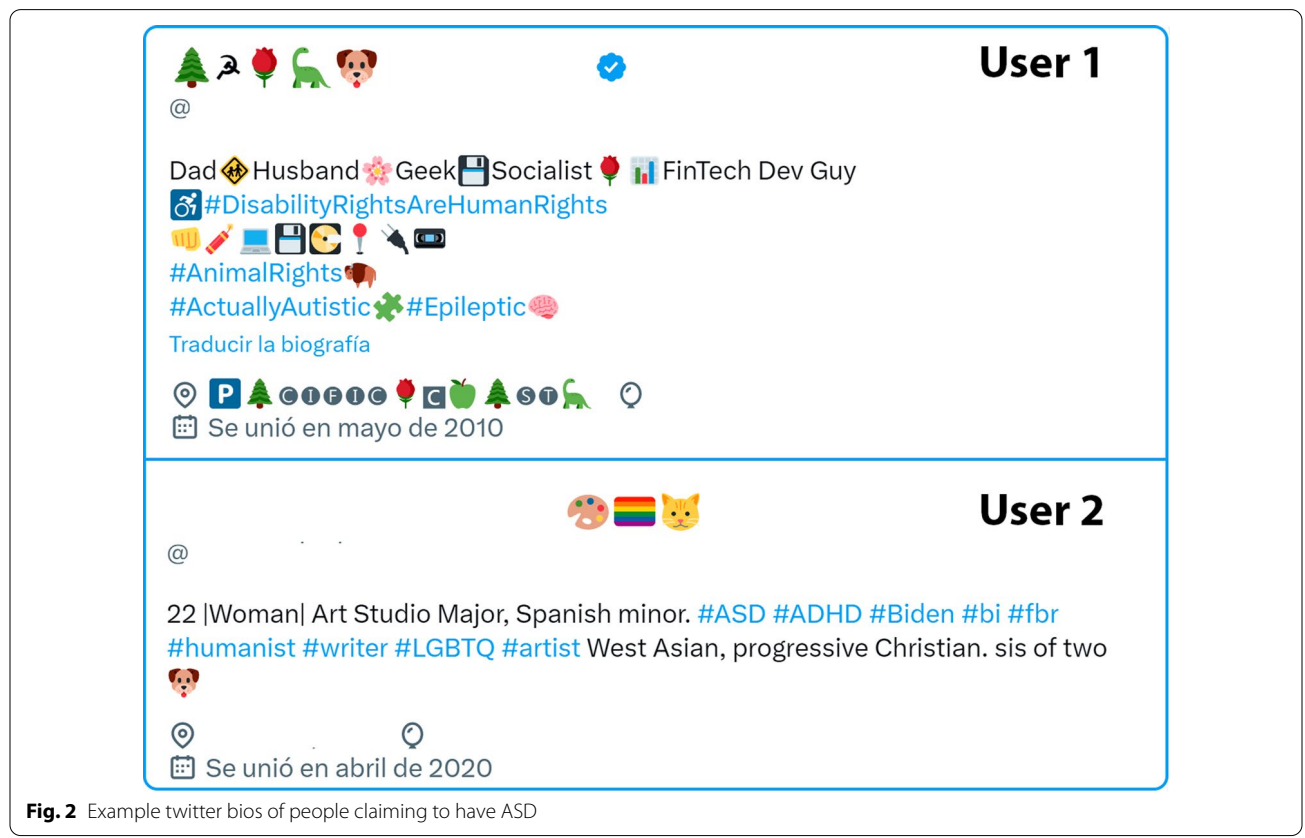

**Fig. 2** Example twitter bios of people claiming to have ASD

**Machine learning and deep learning models used**

An intriguing research avenue was explored, focusing on the evaluation of ML and DL models to identify the most accurate approach for addressing the problem. Consequently, different ML models were employed in this study to obtain results. While these models were tested, there was also a curiosity to investigate the potential of DL models and assess whether they could outperform traditional models in terms of accuracy. The following section provides an explanation of the models utilized:

- Decision trees [23]: A machine learning model that excels in situations where nonlinear relationships among variables are prominent. It provides superior mapping capabilities compared to other models. In this model the decision trees are built using an algorithm that splits recursively the data into several small sequences, which helps to give focus to the important features of the data. This process still be done until a certain stop requirements are achieved, like reaching the depth limit of the tree.
- XGB (eXtreme Gradient Boosting) [24]: This machine learning model is based on random forests (RF) but incorporates several optimizations. It operates by initially considering only a subset of randomly selected variables, repeating this process multiple times with different variables. Additionally, each tree takes into account the results of the previous tree, giving importance to the misclassified instances. After creating each tree, the error is calculated, which helps to create another tree that has to correct that error margin. The combination of trees helps give an accurate prediction because this model implements different techniques to avoid overfitting.
- KNN (K-Nearest neighbours) [25]: A machine learning model that, after being trained, takes into account the K nearest classified values from the testing sample. The result is influenced by its neighboring instances, conditioning the outcome. The most important in this model is to establish a 'k' that helps to get the best accuracy. In this model, 'k' is the number of nearest neighbours that are going to take into account for giving a prediction about in which group should be classified the current data point. In addition, the method or technique used to is vital because exist different alternatives to do this labor, but for this model it is used the computation of euclidean distance. In the prediction realised by this model, it assigns for each data point the most common category found in its 'k' nearest neighbours. This model does not "learn" as it could be known, in fact it just holds a copy of the data used as train and does predictions with the new data.
- RNN (Recurrent Neural Network) [26]: This kind of neural networks uses with an internal state which works as a memory that process sequences of inputs, in this case texts. The key feature of RNNs is the recurrent connections that are established in the network because the previous output or the previous state of the network helps to predict the current prediction. However, the main problem that is found with RNN is the vanishing gradient problem, where the contributions of the context or other information decays geometrically over time which makes the backward propagation through time training of the simple RNN ineffective.
- LSTM (Long Short-Term Memory) [27]: it is a sophisticated type of RNN which overcomes the disadvantage explained before. Unlike standard RNNs. LSTMs keep a cell state over time, which transmits information from earlier steps to later ones. This helps LSTM to keep track of dependencies in the input over longer periods of time. To control the flow of information, LSTM uses three types of gates, all of which are like small neural networks with sigmoid activations. The three types of gates are:

  – Input Gate, which controls how much of the newly calculated state for the current prediction should be stored in the cell state, in other words, filters the information to discard the unimportant information.
  – Forget Gate, which decides what kind of percentage from the previous cell state ought to be kept. It helps to forget irrelevant sections of the information.
  – Output Gate, which controls how much of the internal cell state should be exposed to the next layers of the network.
  These gates perform a multiplication operation with either the input and the previous state to give a better and accurate prediction. Indeed, LSTM and Bi-LSTM have the same underlying structure, the main difference lies in Bi-LSTM, which uses a bidirectional LSTM layer that considers the text sequence not just from left to right but right to left too, taking into account the context of the information.

- BERT (Bidirectional Encoder Representations from Transformers) [28]: A deep learning model that generally achieves high accuracy in natural language processing (NLP) tasks. It is particularly suitable for this study as its encoder reads the complete word sequence from left to right and vice versa, taking into consideration the contextual information of surrounding words. Moreover, BERT is a pre-trained model, which means it has been trained on a large corpus of text data. While the pre-training process, BERT learns to predict missing words in a sentence and to distinguish well structured information from random ones. This model is ideal for cases where named entity recognition or sentiment analysis are done among other activities [29]. So, this model is on the best candidates to give the most accurate percentage of accuracy among the other models. Hence, although for this research is used a BERT-base model, it is also used a variant or an specialised model of BERT which is called BERTweet,which was pre-trained concretely on a corpus of tweets. Tweets usually contains informal language, expressions, emojis and abbreviations that are not commonly found in large amount of web texts. So, BERT-base and BERTweet [30] use the same underlying model architecture but they differ in the type of data that they were pre-trained on.

### Evaluation metrics

To evaluate the models, the results are displayed through confusion matrices. In this way it is possible to visualise the performance of the classification models. The confusion matrix consists of the following elements showed in Table 1.

From the confusion matrix, several performance metrics can be computed:

- Accuracy: The proportion of correct predictions among the total number of cases.

$$\text{Accuracy} = \frac{TP + TN}{TP + TN + FP + FN}$$

- Precision (or Positive Predictive Value): The proportion of positive identifications that were actually correct.

$$\text{Precision} = \frac{TP}{TP + FP}$$

- Recall (or Sensitivity or True Positive Rate): The proportion of actual positives that were correctly identified.

$$\text{Recall} = \frac{TP}{TP + FN}$$

- F1 Score: The harmonic mean of precision and recall.

$$F1 = 2 \times \frac{\text{Precision} \times \text{Recall}}{\text{Precision} + \text{Recall}}$$

- Specificity (or True Negative Rate): The proportion of actual negatives that were correctly identified.

$$\text{Specificity} = \frac{TN}{TN + FP}$$

These metrics provide a comprehensive view of the model's performance, especially in cases where the classes are imbalanced.

### Hardware and Software used for the experiments

To conduct all the experiments, two separate Jupyter Notebooks were employed. Both notebooks utilized Python 3.9 programming language and were executed on a computer with the following specifications: Intel(R) Core(TM) i7-9700K CPU @ 3.60GHZ, 32.0GB RAM, and an NVIDIA GeForce RTX 2080 graphics card.

## Experiments and results

### Data extraction and pre-processing

The initial step involved manually identifying users for the experiments by searching for specific keywords, including 'Autism', 'ASD', 'Asperger', 'Aspie', 'Autistic', and 'ActuallyAutistic', within their biographies. This selection process was carried out diligently, with each profile being manually reviewed. Consequently, several users were excluded as they did not correspond to individuals claiming to have ASD. The following user categories were discarded from the ASD group:

- Profiles belonging to organizations or societies.
- Users who identified themselves as ASD advocates rather than patients.
- Family members of individuals with ASD, such as those who mentioned being the 'Father of an ASD kid', 'Mother of an ASD kid', or part of an 'ASD family'.

Data was obtained from public Twitter users using a Python script programmed to interact with Twitter's API developer. This facilitated the extraction of user publications, which were then exported to a CSV file. The tweets were collected from January 1st, 2017, to January 31th, 2022, covering a period of approximately five years. The dataset was designed to consist of two groups: a representative number of ASD patients and a group of individuals without ASD.

**Table 1** Elements of a confusion matrix

| Actual | Predicted | |
|---|---|---|
| | Positive | Negative |
| Positive | True Positive (TP) | False Negative (FN) |
| Negative | False Positive (FP) | True Negative (TN) |

Subsequently, several pre-processing steps were applied to the data, which are outlined as follows:

- Removal of duplicate tweets or those with identical content.
- Elimination of retweets posted by the users.
- Exclusion of tweets that were not extracted correctly or in their entirety.
- Removal of tweets automatically published by users through sharing options from other platforms like YouTube and Facebook.

For the experiment, a total of 252 users were considered, with 221 classified as ASD users and 31 classified as non-ASD users. Prior to the pre-processing procedure, the dataset consisted of 1,014,723 classified tweets. After undergoing the aforementioned steps, the dataset was reduced and cleaned, resulting in 404,627 tweets. From the complete dataset, a subset of 90,000 tweets was selected with an equal distribution of 45,000 from ASD and non-ASD users respectively.

**Implementation of machine learning and deep learning models**

The dataset was randomly divided into training and testing sets, with 75% allocated for training the models and 25% for testing. The primary objective was to identify the best-performing model and compare their results to determine the most accurate model in this specific context. To achieve optimal results, an investigation of the best hyperparameters, which contribute to improving model performance, was conducted. This process, known as hyperparameter search, was facilitated using the Python library called GridSearchCV.

**Table 2** ML models' hyperparameters

| Model | Parameter | Value |
| --- | --- | --- |
| Decision Trees | Max_depth | 9 |
| | Min_samples_leaf | 2 |
| | Min_samples_split | 4 |
| XGB | Colsample_bytree | 0.8 |
| | Gamma | 5 |
| | Max_depth | 5 |
| | Min_child_weight | 1 |
| | Subsample | 0.8 |
| KNN | N_neighbors | 1 |

The hyperparameters for each ML model are outlined below, in Table 2:

The RNN model is made up of the following layers:

- Embedding layer.
- Simple RNN layer with 64 units.
- Two fully connected layers with dropout between them
- The final output layer has a single neuron due to the fact that is the responsable for classifying the sample.

In Fig. 3 the scheme of the RNN arquitecture is shown.
The LSTM model is made up of the following layers:

- The input pass through a process of text vectorization.
- Embedding layer.
- LSTM layer with 64 units.
- One fully connected layer.
- The final output layer has a single neuron because is in charge of classifying the sample.

The only difference among the LSTM and Bi-LSTM arquitectures is the LSTM and Bi-LSTM layers. In Fig. 4 the schemes of the LSTM and Bi-LSTM arquitectures are shown.

**Results**

Three ML models, namely decision trees, XGB, and KNN, were trained, alongside other DL models, namely RNN, LSTM, Bi-LSTM, BERT and BERTweet. The results, displayed in Table 3, support the hypothesis that some DL basic models achieves higher accuracy compared to the ML models with hyperparameters.

Figure 5 displays the confusion matrices for eight different classification models utilized in a binary classification task aimed at identifying individuals with Autism Spectrum Disorder (ASD).

The BERTweet model stands out as the top-performing model, exhibiting a significant number of true positives and true negatives, indicating its strong ability to accurately identify individuals with and without ASD. As a deep learning model, BERT leverages neural networks to discern intricate patterns within the input data. This highlights the potential of deep learning models in extracting relevant patterns, thus enhancing the precision of classification.

While hyperparameter optimization was performed for the machine learning models, it was found that the BERTweet model outperformed the others. The KNN model achieved the lowest accuracy at 60.8%, followed by the decision tree with 61.2%, LSTM with 69.5%, RNN with 69.9%, and Bi-LSTM and XGB with an accuracy of 70.3% and 71.6% respectively. Notably, the BERT-based models achieved the best accuracies. The accuracy of BERT and BERTweet models were 84.3% and 87.7 respectively. So the model with the best accuracy was BERTweet.

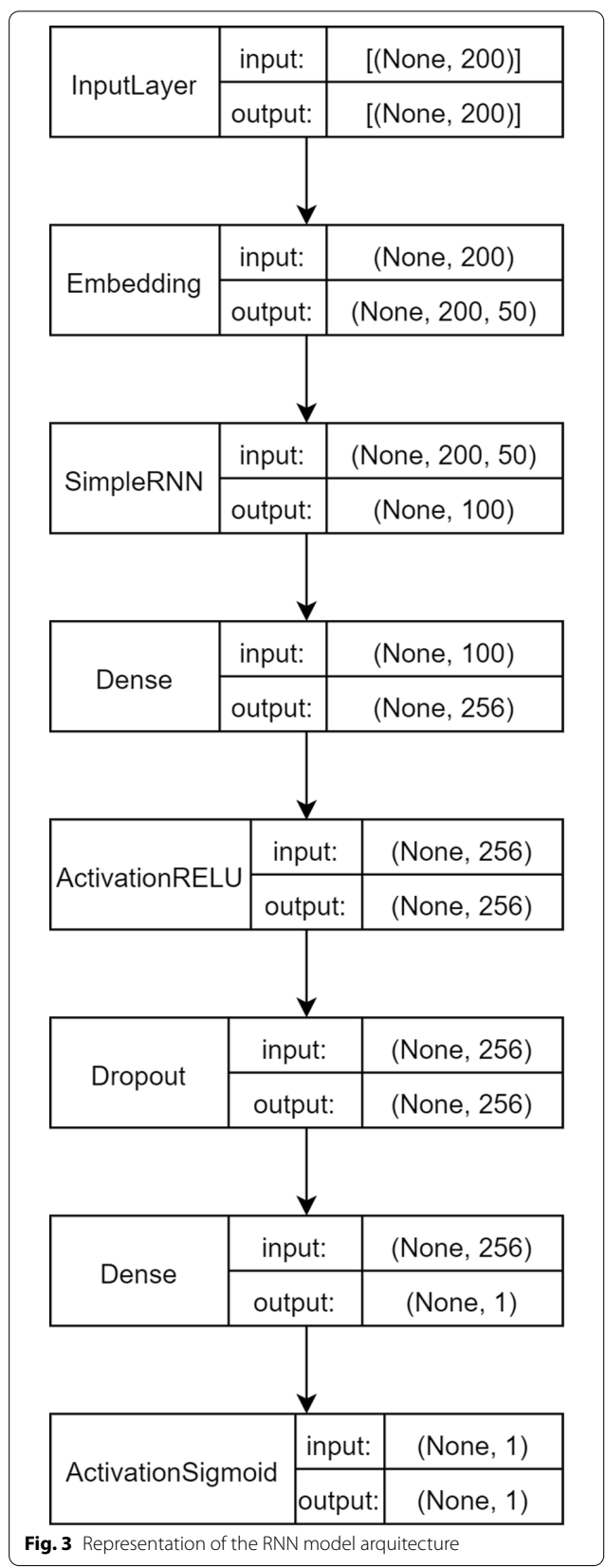

**Fig. 3** Representation of the RNN model arquitecture

In summary, the analysis of the confusion matrices emphasizes the importance of selecting the appropriate model for detecting ASD and evaluating its performance using metrics such as confusion matrices. The exceptional accuracy and ability of the BERT model to learn complex patterns in the data suggest that deep learning models have the potential to significantly enhance the accuracy of classification tasks involving individuals with and without ASD.

**Discussion and conclusions**

In this study, a cohort of Twitter users was examined and classified into two groups: ASD users and non-ASD users. This classification process involved an initial search for specific terms within the users' biographies, followed by a manual review of the selected users' timelines based on their biographical descriptions. The Twitter API was then utilized to collect the users' posts, automatically labeling the texts as originating from either ASD or non-ASD users. The main objective of this research was to develop highly accurate models capable of predicting whether a given text was authored by an ASD user.

After preprocessing the dataset, a subset of 45,000 texts was selected from each group (ASD and non-ASD users), resulting in a total of 90,000 tweets. These tweets were further divided into training and test sets to train various models using both traditional machine learning and deep learning techniques.

As the best ML model it is found XGB with an accuracy of 71.6%. However from the whole number of models that were tested, the best one was the DL model called BERTweet with an accuracy of 87.7%. This aligns with previous studies that have shown that BERT-based models have a great effectiveness in categorizing any kind of text texts, including tweets [18, 31, 32].

It is important to acknowledge certain limitations of this study. One significant limitation is the potential biases in the collected dataset. These biases could arise from users providing false information in their biographies or tweets being authored by individuals other than the profile owners. Moreover, while our models show promise, they are not intended to replace medical specialists. Instead, they aim to assist in identifying potential ASD traits. Before these models can be considered for use in a medical consultation platform, these limitations

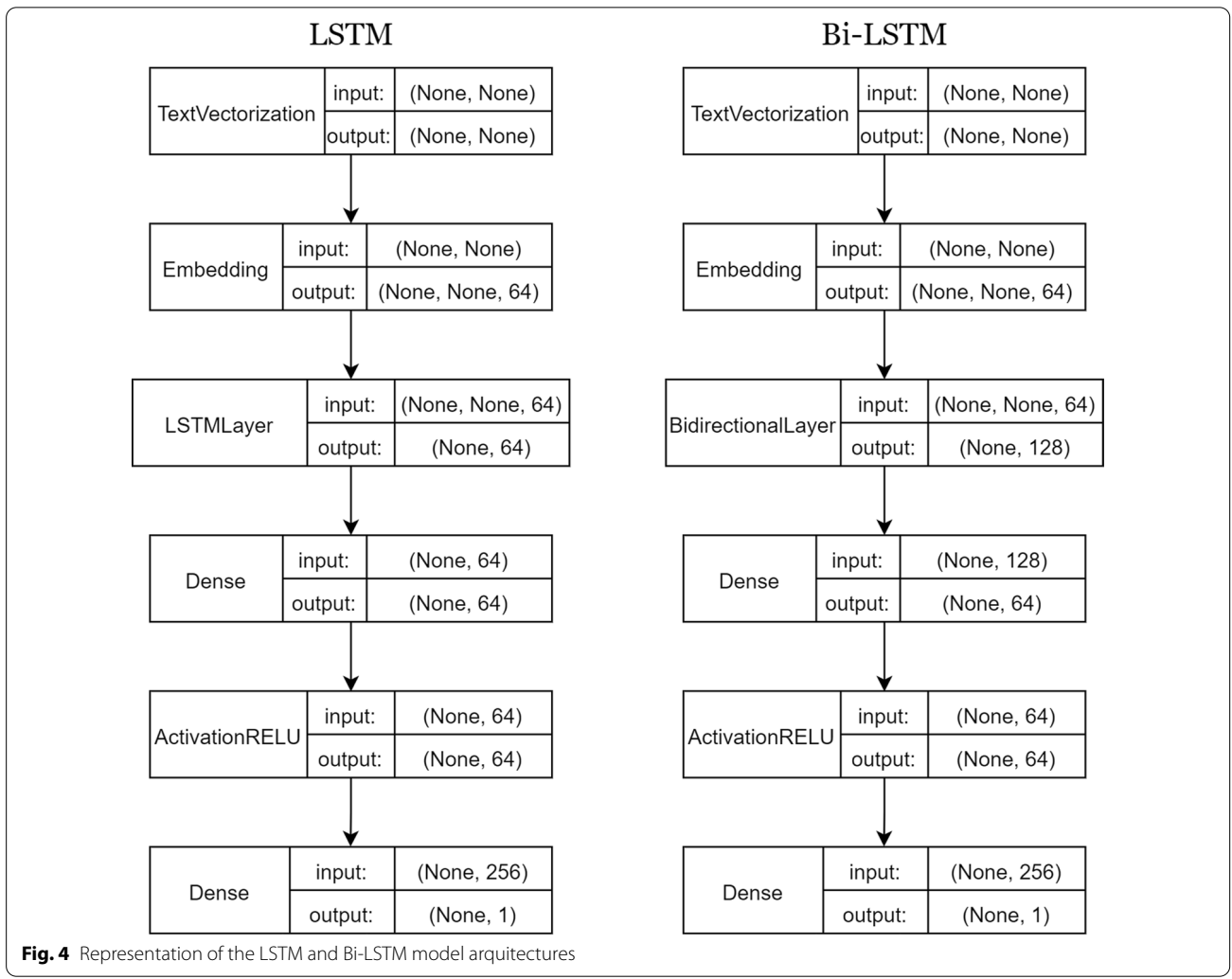

**Fig. 4** Representation of the LSTM and Bi-LSTM model arquitectures

**Table 3** Results of ML and DL models

| Model name | tp | fp | tn | fn | f1 | Acc (%) |
| --- | --- | --- | --- | --- | --- | --- |
| Decision Trees | 11,229 | 8690 | 2540 | 41 | 0.72 | 61.2 |
| XGB | 8895 | 4007 | 7223 | 2375 | 0.74 | 71.6 |
| KNN | 8867 | 6411 | 4819 | 2403 | 0.67 | 60.8 |
| RNN | 8544 | 4050 | 7180 | 2726 | 0.71 | 69.9 |
| LSTM | 4540 | 984 | 10,246 | 6730 | 0.65 | 69.5 |
| Bi-LSTM | 5223 | 1299 | 9931 | 6047 | 0.66 | 70.3 |
| BERT | 9252 | 1515 | 9715 | 2018 | 0.84 | 84.3 |
| BERTweet | 9564 | 1053 | 10,177 | 1706 | **0.88** | **87.7%** |

*tp* true positives, *fp* false positives, *tn* true negatives, *fn* false negatives, *f1* f1-score, *acc* accuracy)

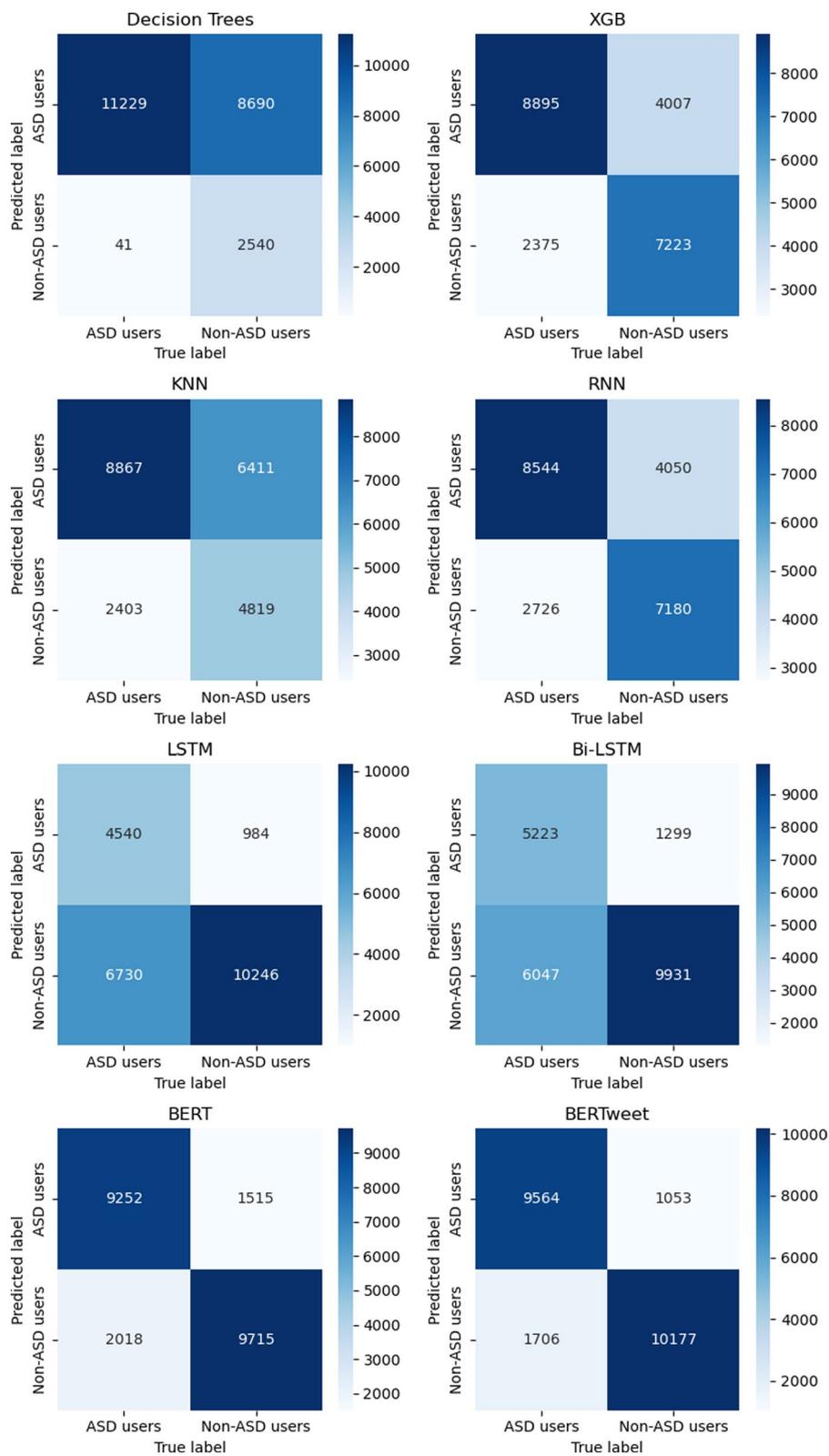

**Fig. 5** Confusion matrices of the 8 trained models (Decision Trees, XGB, KNN, RNN, LSTM, Bi-LSTM, BERT & BERTweet)

must be addressed. However, this research serves as a foundation for future investigations. This includes exploring hyperparameters to further enhance the accuracy of BERTs models and other deep learning models. Additionally, future efforts will involve training and evaluating additional deep learning and machine learning models that have not been previously examined, ensuring that no high-performing models are overlooked.

**Author contributions**
SR-M: Conceptualization, data curation, methodology, software, visualization, validation, writing—original draft preparation. MTG-O: Conceptualization, supervision, writing—reviewing and editing. MB-G: Data curation, writing—original draft preparation. NP-F: Conceptualization, supervision, writing—reviewing and editing. JAB-A: Conceptualization, data curation, methodology, software, visualization, validation, writing—reviewing and editing.

**Funding**
This study receives partial support from Universidad de León through the "Programa Propio de Investigación de la Universidad de León 2021" grant.

**Data availability**
All the data and code used in the experiments, are available in a Github https://github.com/srubm/Dataset-ASD-twitter-users.

**Declarations**

**Conflict of interest**
The authors declare that they have no known competing financial interests or personal relationships that could have appeared to influence the work reported in this paper.

**Author details**
[1]SALBIS Research Group, Dept. of Electric, Systems and Automatics Engineering, Universidad de León, Campus of Vegazana s/n, 24071 León, León, Spain. [2]SECOMUCI Research Group, Dept. of Electric, Systems and Automatics Engineering, Universidad de León, Campus of Vegazana s/n, 24071 León, León, Spain.

**References**
1. ASD—what is autism spectrum disorder? Pediatric Patient Education. 2023 Mar; eprint: https://publications.aap.org/patiented/article-pdf/doi/10.1542/ppe_document143/1462374/ppe_document143_en.pdf.
2. Salari N, Rasoulpoor S, Rasoulpoor S, Shohaimi S, Jafarpour S, Abdoli N, et al. The global prevalence of autism spectrum disorder: a comprehensive systematic review and meta-analysis. Ital J Pediatr. 2022;48(1):112. https://doi.org/10.1186/s13052-022-01310-w.
3. Hosseini SA, Molla M. Asperger syndrome. Treasure Island: StatPearls Publishing; 2022.
4. Fernell E, Eriksson MA, Gillberg C. Early diagnosis of autism and impact on prognosis: a narrative review. Clin Epidemiol. 2013;5:33–43. https://doi.org/10.2147/CLEP.S41714.
5. Gabbay-Dizdar N, Ilan M, Meiri G, Faroy M, Michaelovski A, Flusser H, et al. Early diagnosis of autism in the community is associated with marked improvement in social symptoms within 1–2 years. Autism. 2022;26(6):1353–63. https://doi.org/10.1177/13623613211049011.
6. Zwaigenbaum L, Bauman ML, Choueiri R, Kasari C, Carter A, Granpeesheh D, et al. Early intervention for children with autism spectrum disorder under 3 years of age: recommendations for practice and research. Pediatrics. 2015;136(Supplement 1):S60-81. https://doi.org/10.1542/peds.2014-3667E.
7. Anderson DK, Oti R, Lord C, Welch K. Patterns of growth in adaptive social abilities among children with autism spectrum disorders. J Abnorm Child Psychol. 2009;37:1019–34.
8. Barger B, Campbell JM, McDonough JD. Prevalence and onset of regression within autism spectrum disorders: a meta-analytic review. J Autism Dev Disord. 2013;43:817–28.
9. Woolfenden S, Sarkozy V, Ridley GF, Williams KJ. A systematic review of the diagnostic stability of autism spectrum disorder. Res Autism Spectr Disord. 2012;6:345–54.
10. Hernández-Chan GS, Ceh-Varela EE, Sanchez-Cervantes JL, Villanueva-Escalante M, Rodríguez-González A, Pérez-Gallardo Y. Collective intelligence in medical diagnosis systems: a case study. Comput Biol Med. 2016;74:45–53. https://doi.org/10.1016/j.compbiomed.2016.04.016.
11. Najafabadipour M, Zanin M, Rodríguez-González A, Torrente M, Nuñez García B, Cruz Bermudez JL, et al. Reconstructing the patient's natural history from electronic health records. Artif Intell Med. 2020;105: 101860. https://doi.org/10.1016/j.artmed.2020.101860.
12. Goldenberg SL, Nir G, Salcudean SE. A new era: artificial intelligence and machine learning in prostate cancer. Nat Rev Urol. 2019;16(7):391–403. https://doi.org/10.1038/s41585-019-0193-3.
13. Islam MM, Haque MR, Iqbal H, Hasan MM, Hasan M, Kabir MN. Breast cancer prediction: a comparative study using machine learning techniques. SN Comput Sci. 2020;1(5):290. https://doi.org/10.1007/s42979-020-00305-w.
14. Mehmood M, Rizwan M, Ml MG, Abbas S. Machine learning assisted cervical cancer detection. Front Public Health. 2021;9: 788376.
15. Eslami T, Mirjalili V, Fong A, Laird AR, Saeed F. ASD-DiagNet: A hybrid learning approach for detection of autism spectrum disorder using fMRI data. Front Neuroinform. 2019;13. 10.3389/fninf.2019.00070.
16. Nogay HS, Adeli H. Machine learning (ML) for the diagnosis of autism spectrum disorder (ASD) using brain imaging. Rev Neurosci. 2020;31(8):825–41. https://doi.org/10.1515/revneuro-2020-0043.
17. Carette R, Elbattah M, Dequen G, Guérin JL, Cilia F. Visualization of eye-tracking patterns in autism spectrum disorder: method and dataset. In: 2018 Thirteenth international conference on digital information management (ICDIM); 2018. p. 248–53.
18. Benítez-Andrades JA, Alija-Pérez JM, Vidal ME, Pastor-Vargas R, García-Ordás MT. Traditional machine learning models and bidirectional encoder representations from transformer (BERT)-based automatic classification of tweets about eating disorders: algorithm development and validation study. JMIR Med Inform. 10(2): e34492. https://doi.org/10.2196/34492.
19. Bullard J, Ovesdotter Alm C, Liu X, Yu Q, Proaño R. Towards early dementia detection: fusing linguistic and non-linguistic clinical data. In: Proceedings of the Third workshop on computational linguistics and clinical psychology. San Diego, CA, USA: Association for Computational Linguistics; 2016. p. 12–22. Available from: https://aclanthology.org/W16-0302.
20. Batsakis S, Adamou M, Tachmazidis I, Jones S, Titarenko S, Antoniou G, et al. Data-driven decision support for adult autism diagnosis using machine learning. Digital. 2022;2(2):224–43. https://doi.org/10.3390/digital2020014.
21. Thelwall S, Thelwall M.: Autism spectrum disorder on Twitter during COVID-19: account types, self-descriptions and tweeting themes [SSRN Scholarly Paper]. Rochester, NY. Available from: https://papers.ssrn.com/abstract=3826169.
22. Rubio-Martín S, García-Ordás MT, Bayón-Gutiérrez M, Prieto-Fernández N, Benítez-Andrades JA. Early detection of autism spectrum disorder through AI-powered analysis of social media texts. In: 2023 IEEE 36th International symposium on computer-based medical systems (CBMS); 2023. p. 235–240.
23. Quinlan JR. Induction of decision trees. Mach Learn. 1986;1(1):81–106. https://doi.org/10.1007/BF00116251.
24. Chen T, Guestrin C. XGBoost: A scalable tree boosting system. In: Proceedings of the 22nd ACM SIGKDD International conference on knowledge discovery and data mining. KDD '16. New York, NY, USA: Association for Computing Machinery; 2016. p. 785-794. Available from: https://doi.org/10.1145/2939672.2939785.
25. Beckmann M, Ebecken NFF, de Lima BSLP. A KNN undersampling approach for data balancing. J Intell Learn Syst Appl. 2015;7:104–16.
26. Hopfield J. Neural networks and physical systems with emergent collective computational abilities. Proc Natl Acad Sci USA. 1982;05(79):2554–8. https://doi.org/10.1073/pnas.79.8.2554.
27. Hochreiter S, Schmidhuber J. Long short-term memory. Neural Comput. 1997 11;9(8):1735–1780. https://doi.org/10.1162/neco.1997.9.8.1735.

28. Devlin J, Chang MW, Lee K, Toutanova K. BERT: pre-training of deep bidirectional transformers for language understanding. In: Proceedings of the 2019 Conference of the North American Chapter of the Association for Computational Linguistics: Human Language Technologies, Volume 1 (Long and Short Papers). Minneapolis, Minnesota: Association for Computational Linguistics; 2019. p. 4171–4186. Available from: https://aclanthology.org/N19-1423.
29. Benítez-Andrades JA, González-Jiménez A, López-Brea A, Aveleira-Mata J, Alija-Pérez JM, García-Ordás MT. Detecting racism and xenophobia using deep learning models on Twitter data: CNN. LSTM and BERT PeerJ Comput Sci. 2022;8: e906. https://doi.org/10.7717/peerj-cs.906.
30. Nguyen DQ, Vu T, Nguyen AT. BERTweet: a pre-trained language model for English Tweets. In: Proceedings of the 2020 Conference on empirical methods in natural language processing: system demonstrations; 2020. p. 9–14.
31. Yu S, Su J, Luo D. Improving BERT-based text classification with auxiliary sentence and domain knowledge. IEEE Access. 2019;7:176600–12. https://doi.org/10.1109/ACCESS.2019.2953990.
32. Palanivinayagam A, El-Bayeh CZ, Damaševičius R. Twenty years of machine-learning-based text classification: a systematic review. Algorithms. 2023. https://doi.org/10.3390/a16050236.